# Intelligent Biohybrid Neurotechnologies: Are They Really What They Claim?

Gabriella Panuccio[1*], Marianna Semprini[1], Lorenzo Natale[2], Michela Chiappalone[1]

[1]Dept of Neuroscience & Brain Technology, Neural Interface & NEurorehabilitation (NINE) Lab, Istituto Italiano di Technologia, Genova, Italy; [2]iCub Facility, Istituto Italiano di Technologia, Genova

\* Correspondence to:  **Gabriella Panuccio, M.D., Ph.D.**
Istituto Italiano di Technologia
Dept of Neuroscience & Brain Technologies – NINE Lab

Via Morego, 30
16163 Genova
ITALY

email: gabriella.panuccio@iit.it
telephone: +39 010 71781 580
cell phone: +39 333 491 32 31

## Graphical abstract

In the era of 'intelligent' biohybrid neurotechnologies for brain repair, new fanciful terms are appearing in the scientific dictionary to define what has so far been unimaginable. As the emerging neurotechnologies are becoming increasingly polyhedral and sophisticated, should we talk about evolution and rank the intelligence of these devices?

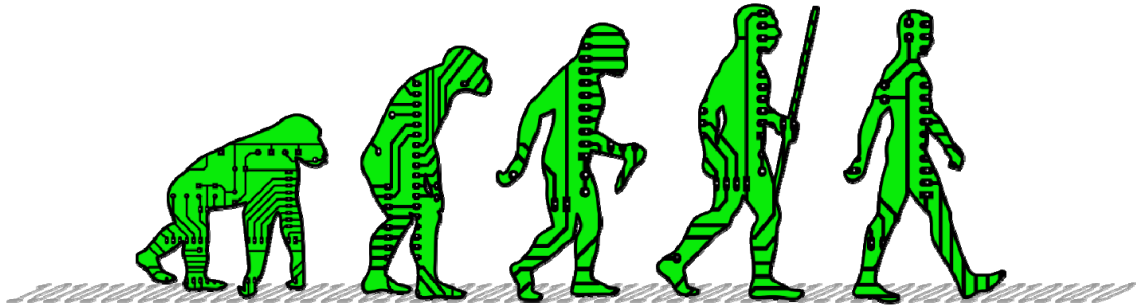

## Introduction

In the era of visionary neurotechnologies at the edge between biology and engineering, 'intelligent' biohybrid systems have recently come into play. These systems are based on the interaction between a biological component (here, the nervous tissue) and an artificial device whose interaction is mediated by intelligent control algorithms exhibiting different levels of sophistication.

Intelligent biohybrid systems have become increasingly popular within the brain repair scientific community, up to the point of representing today the leading unconventional brain repair strategy. However, the variety of acceptations as to what *hybrid* or *intelligent* means is leading to different understandings of how a device of biohybrid architecture and intelligent operation should be conceived, with significant implications for the bioengineering, clinical and industrial settings. Indeed, the terms *biohybrid* and *intelligent* are liberally used, alone or in combination, to describe whatever biological-to-artificial interface the human imagination may suggest. To add to the confusion, the cross-disciplinary approach at the core of biohybrid devices has lead to the cross-contamination of commonly used terminology.

Here, we provide clarifying definitions of these terms and we propose a new hierarchical classification of biohybrid systems for the neurotechnologies field, based on combined architecture and feature criteria, with the intent to terminate the untamed use of colorful terminology and bring up a unifying scheme to define such innovative systems that are leading the era of implantable devices for brain repair.

## What is a 'hybrid' system? And do we need that 'bio' by the way?

Everyone in the neurotechnologies field has certainly studied the revolutionary theories of Ludvig von Bertalanffy[1] and surely knows how to define a system. But what is a *hybrid* system? A hybrid system is generated whenever the system's components are of different nature. A straightforward example of everyday-life is a vehicle engine that uses two different (alternative) sources of energy (e.g., fuel and electricity), i.e., the much *en vogue* hybrid car (Fig. 1a).

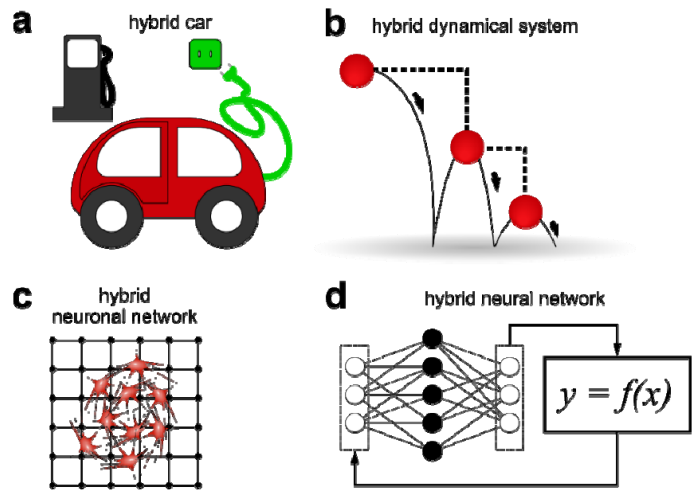

**Figure 1 – Hybrid systems**
**(a)** A hybrid car engine is powered by fuel or by electricity. **(b)** A hybrid dynamical system is described both in the continuous and the discrete domains. For example, a ball bouncing off the ground describes a continuous trajectory (solid line) while losing energy and thus decreasing its speed in a step-like (discrete) fashion (dashed line) after each collision with the ground. **(c)** Hybrid neuronal networks are made of biological neurons and artificial neural models. **(d)** Hybrid neural networks are made of model neurons and symbolic functions.

When at least one of the system's components is of biological origin, the system *should* be regarded as 'bio'-hybrid. However, this is not always the case. Indeed, biomedical engineers (and especially those working in the field of neurotechnologies) have started dominating the concept of hybrid systems by giving for granted that hybrid is anything that contains something that is not artificial (namely, biology). However, from a biologist's or a health-care professional's perspective, a hybrid system is anything that contains something that is neither biological nor human (namely, artificial). For dynamical systems theorists a hybrid system is even another totally different business (Fig. 1b). More pertinent to the here discussed topic are hybrid networks populated by biological and artificial neurons. Going back to the stated objection on misused terminology, a hybrid *neuronal* network is made up by biological neurons interacting with artificial neural models (Fig. 1c), whereas a hybrid *neural* network refers to a completely artificial hybrid system, where the difference in the components' nature refers to the description formalism: for example, biophysically detailed computational models of neurons may be coupled to symbolic functions providing an arbitrary mathematical relationship among them (Fig. 1d).

The examples to bring about in support of our point are numerous, but these few considerations are compelling enough to conclude that the term *hybrid* is by itself rather unspecific and not informative about the characteristics of a system's components. On the other hand, the term 'bio'-hybrid specifies that the system is hybrid *and* that it is made up by biological and artificial elements. Whereas this may be generally agreed upon in the case of macroscopic evident robotic end effectors connected to biological nervous tissue (Fig. 2a), it is not as much common in the case of intracranial electronic devices, such as a DBS apparatus (Fig. 2b).

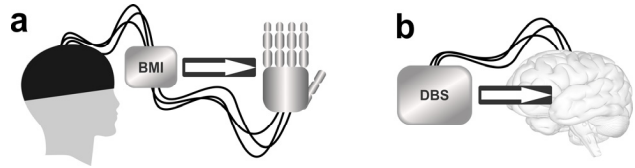

**Figure 2 – Biohybrid systems in the neurotechnologies: two practical examples**
**(a)** A BMI that conveys the electrical activity of the recorded brain area to the robotic end effector establishes a biohybrid system made up by the input brain area, the interface apparatus and the robotic hand. **(b)** A DBS device influences the electrical activity of the stimulated brain area without necessarily receiving a feedback message from it. By virtue of the influence exerted by the DBS device on the stimulated brain area, these two elements constitute a biohybrid system.

## What is an 'intelligent' system?

The foundation of artificial intelligence (AI) as an academic discipline dates back to the Dartmouth Conference (now regarded as 'the birth of AI'[2]) in 1956, which represented the culmination of a series of world-changing events contributed by visionary scientists committed to build *humane* machines. In the wake of this legacy, the implementation of AI to the treatment of neurological disorders has revolutionized the concept of brain repair strategies. However, defining a system according to its components is straightforward, whereas classifying a system's intelligence is not. Indeed, such classification would imply 'scoring' the device's capabilities, which is not trivial. The literature itself is also quite obscure on this point, because virtually any system receiving an input and responding to it is nowadays regarded as 'intelligent'. We therefore need to go back to the definition and meaning of intelligent system as provided by the pioneers of AI.

The invention of the first general-purpose programmable digital computer, the Z3 by Konrad Zuse in 1941[3], inspired the vision of building an electronic brain, which immediately reminds of Isaac Asimov's "Robot Series", a collection of globally recognized 'sci-fi' novels on positronic-brain robots. In 1948 Norbert Wiener described "computing machines" (i.e. computers) that were able to improve their behavior during a chess competition by analyzing past performances[4]. These machines may be regarded as self-evolving: they are driven by a feedback mechanism based on the evaluation of previous failures/successes to adjust their behavior in response to experience and performance scoring. In the same years, Grey Walter built the turtle robots, the first electronic autonomous robots[5,6]. He wanted to prove that rich connections between a small numbers of brain cells could give rise to very complex behaviors such as making decisions. He demonstrated that a system composed of few elements produced unpredictable behavior, which he defined as "free will". These extraordinary science achievements wiped out traditional certainty about machines, as set down one century before by Lady Ada Lovelace who stated that "machines only do what we tell them to do" (i.e. they do not evolve)[7]. This statement was formally rejected by Alan Turing in his seminal paper "Computing Machinery and Intelligence"[8], in which he argued that the objection could be proven wrong by building a *learning machine*. He also explained that this was only possible if instead of building a machine simulating the adult brain, we built a *child machine* having some *hereditary material* (predefined structure), subject to *mutation* (changes) and *natural selection* (judgment of the experimenter) and undergoing a learning process based on a reward/punishment mechanism (best examples are today's learning systems based on deep-reinforcement learning[9]). Turing therefore defined *intelligence* as the ability to reach human levels of performance in a cognitive task (Turing Test). In robotics, the field of *embodied intelligence* challenges the conventional view that sees intelligence as the ability to manipulate symbols and produce other symbols. *Embodied intelligence* focuses instead on the skills that allow biological systems to thrive in their environment: the ability to walk, catch prays, escape from threats, and, above all, *adapt*.

From this brief *excursus* it emerges that *autonomous behavior* (Walter's "free will"), intended as the agent's ability to make a decision in response to an external event, and *evolution*, intended as the agent's ability of adapting to changes in the environmental stimuli so to maximize a given reward, are both attributes of intelligence, as they are both appropriate responses to external stimuli. In fact, the term intelligent originates from the Latin verb *intelligere,* the union of *inter* (between) and *legere* (read), meaning *read between the lines*, *comprehend*, *understand*, which makes a substantial difference between reading or noting something and understanding the meaning of it in order to take consequent action.

According to this definition, the feature of detecting an electrical event generated by the brain without interpreting its meaning should not suffice for a system to earn the recognition of intelligence, whereas *autonomous behavior* and *evolution* should be the appropriate prerequisites, because they imply the capability of interpreting a brain signal (the *environment variable*) so to choose the most appropriate action (*thrive in the environment*). However, as already emphasized, it seems that in the modern acceptance these are no longer necessary conditions, since any system just executing a predefined and stereotyped task in response to a detected event is often advertised as 'intelligent' (we add, although maybe not brilliant).

*Is then reciprocal (bidirectional) communication between biological and artificial elements a sufficient feature to define a biohybrid system as intelligent?* Given the modern acceptance of bidirectional communication, we doubt. The scientific community has converged to the general consensus that establishment of a functional partnership between biological and artificial components *via* bidirectional communication is a crucial prerequisite to successfully achieve functional brain repair. Bidirectional communication implies that a signal generated by the brain can be detected by the artificial device, which in turn takes actions that influence brain activity, which is again forwarded back to the artificial device. This paradigm is regarded as *closed-loop*, as opposed to *open-loop* designs, where the artificial device just 'goes by itself' (Fig. 3).

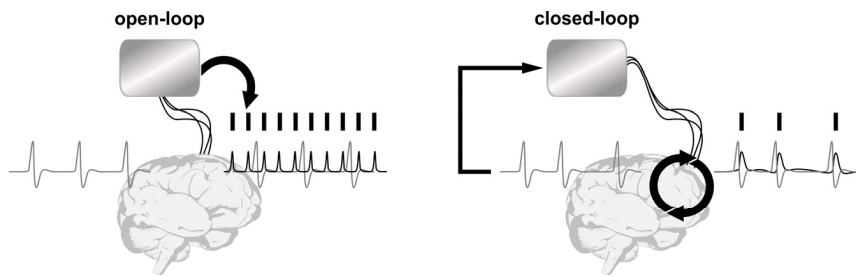

**Figure 3 – Open-loop and closed-loop paradigms**
Depicted are two types of neuromodulators interfering with brain activity by electrical pulses. The open-loop system cannot read brain electrical activity and thus operates independent of it. The communication loop between the brain and the artificial device stays open due to the lack of feedback from the brain. The closed-loop device reads the electrical activity of the brain and operates according to it. The feedback provided by the brain activity closes the communication loop between the brain and the artificial device. Grey trace: ongoing electrical brain activity. Black trace: evoked brain electrical response. Black vertical lines: output pattern of the neuromodulator.

As the closed-loop paradigm is generally regarded as intelligent *per se*, the progressive parallelism between closed-loop operation and intelligent behavior might be the root of all evil. In fact, in most of closed-loop systems the artificial component is not set to autonomously evolve and adapt to the biological counterpart in order to achieve the desired functional outcome. Rather, it responds in a quite stereotyped fashion to the biological inputs it receives. For example, some devices may provide activity-dependent stimulation, i.e., an electrical pulse of predefined parameters is delivered after a predefined delay following the detection of an electrographic event, i.e., stimulation is phase-locked to brain activity[10, 11]. A more advanced paradigm is represented by stimulating devices with built-in knowledge of the best stimulation policy to be implemented depending on the type of input signal[12]. The operation of these recent 'intelligent' devices follows the electrical activity of the brain. In turn, the brain pattern changes over time because it is being modulated, thus influencing the electrical output of the stimulating device. Nonetheless, neither of these (commonly defined intelligent) devices can *autonomously* change and *adapt* their stimulation policy to the progression of the electrical activity generated by the stimulated brain area. The detected electrical signals sustain a rather constrained pseudo-adaptive behavior in that (1) the electrical rhythm of the input brain area dictates the stimuli distribution and (2) the choice of a different output relies on built-in policies. In practice, the machine is instructed to provide OUTPUT-X if INPUT-X happens. Outstanding advancements have been recently made leading to self-adjusting neuroprostheses[13], adaptive algorithms for DBS technology to treat epileptic disorders[14, 15], the intelligent wheelchair[16] and the nurse robot (*nursebot*)[17]. Based on statistical machine learning techniques, these algorithms exhibit autonomous decisional power and evolution, overcoming the initial expectations of the designers as they are capable of elaborating the best (mostly unexpected) intervention policy to achieve a specified goal, rather than picking the predefined most appropriate task. Undoubtedly, as compared to their bidirectional fellows, these are top-class intelligent devices. Thus, bidirectional communication between biological tissue and an artificial device is necessary but definitely not sufficient to endow a biohybrid system with *true* intelligent performance.

## Spring-cleaning the field of biohybrid systems & its misplaced terminology

From what so far argued, it clearly shows that there is much confusion as to what the exact terminology to describe intelligent biohybrid neurotechnologies should be. It is indeed not uncommon to encounter different terms describing the same concept or different concepts explained by the same technical terms. Subtle differences in the operating mode and in the performance of such technologies often lead to the generation of fancy epithets that only kindle the commotion of an already chaotic technical dictionary. These catchy virtuosities are at times misleading for the experts themselves, so surely they do not help those outside the field to immediately grasp the *quid* and thus the translational relevance of the described neurotechnology. However, in the era of converging sciences where the bench-to-industry-to-bedside approach is the *leitmotiv* of global progression, it is imperative to build a common language that would be understandable by any actor.

In order to definitely clarify these issues, we bring back to light the canonical formalism proposed decades ago by Russel and Norving[2]:

> "**An agent is** anything that can be viewed as **perceiving** its environment through **sensors** **and acting** upon that environment through **effectors.** [...]
> A **rational agent** is one that does the right thing. [...] an ideal rational agent should do whatever action is expected to maximize its performance measure. [...]
> A system is **autonomous** to the extent that its behavior is determined by its own experience. [...]A truly autonomous intelligent agent should be able to operate successfully in a wide variety of environments, given sufficient time to adapt".

We thus distinguish in the first place between an *agent* and a *rational agent*, the latter being capable of *autonomous* (human-independent) behavior. Translating these canonical concepts to the design of intelligent biohybrid systems, the intelligent artificial component of the conceived devices should exhibit the capability of dynamic adaption to the flow of neuronal information that is continuously changing due to the reciprocal interaction between artificial and biological components making up the biohybrid system. That is, a truly intelligent system should intrinsically exhibit: **(1)** ability to acquire information (*can read input*), **(2)** set of choices (*have options*) and **(3)** autonomous decisional power (*decide by itself*). Only the synergetic enforcement of these three salient features will allow a system to accomplish its task in an intelligent manner, achieving rationality in its performance independent of human intervention. Such intelligent behavior implies the pivotal role of a learning process based on feedback mechanisms, cost and reward functions.

## A unifying framework for intelligent systems and some practical examples

We formalize the considerations discussed so far by providing a pragmatic[18] solution. We refer to Russell and Norvig[2] to rank the intelligence of a system, based on a hierarchical scheme. We apply these concepts to the algorithms designed to establish a functional dialogue between man and artificial device (e.g., a brain stimulator or a robotic limb) for therapeutic purposes, keeping them distinguished from their implementation in intelligent assistive technologies, such as *humanoid robotics*.

We define systems that operate according to an input/output (I/O) function and we use the logic underlying that function (hereafter, the *algorithm*) to classify the system's performance, from the lowest to the highest sophistication, as follows (see Fig. 4):

*Reactive* – The algorithm is based on a single input-single output paradigm. This is a basic I/O system. A typical example is activity-dependent neuromodulation: an electrical pulse or a train of electrical stimuli may be delivered with a fixed delay upon detection of an electrographic event in the recorded brain area, i.e., stimulation is phase-locked to brain activity[10, 11, 19]. Other striking examples are BMIs and neural prostheses, which, in most cases, serve as a unidirectional replacement of the compromised or lost brain function. These devices act as electrographic activity readers and react to the detection of specific electrographic patterns with the activation of an external artificial device, aimed at achieving a motor task[20]. The output parameters are fixed by the human operator, but the output is triggered by an input signal detected by the system. It needs to be noted that this operating mode is often improperly regarded as 'adaptive'. However, this algorithm represents the simplest implementation of control since it does not operate any *autonomous* choice, but it is simply instructed to *react* with a stereotyped output to a detected (input) event. Hence, we regard it as *non* intelligent algorithm (Fig. 5).

*Responsive* – The algorithm may process several inputs and respond with different outputs, but its operation is task-oriented, contextual and conditional. The algorithm is indeed provided with multiple built-in output options and cannot make *autonomous* decisions. For example, a stimulation algorithm may be instructed to modify its output frequency proportionally to the frequency of the recorded event(s) as in Beverlin & Netoff, 2013[12]. Since the human operator dictates the *how to*, the algorithm exhibits a 'passive enforcement behavior' based on what is referred to as *supervised learning* in the machine learning field. We regard this type of agent as *non* intelligent (Fig. 5). A well-known clinical application of this design is the Responsive NeuroStimulation (RNS) system NeuroPace®, which was recently approved by the Food and Drug Administration (FDA) agency as adjunctive DBS therapy for drug-refractory epileptic patients[21].

*Adaptive* – The algorithm implements a multiple input-adaptive-output paradigm, being capable of real-time self-adjustment according to past experience and performance evaluation exploiting reward functions. The algorithm *evolves* in time using the acquired knowledge, exhibiting goal-directed behavior rather than obeying predefined rules, as in[15]. We call this 'intentional behavior algorithm', as the human operator only provides the knowledge of *what* to achieve, whereas the algorithm autonomously chooses the *how to*, exhibiting the highest sophistication. In the machine learning field this paradigm is regarded as *reinforcement learning*[9] as it is based upon a learning problem where only the ultimate goal is known to the machine, while the human operator does not provide any prior knowledge on the *how to*. This *is* an *intelligent* algorithm (Fig. 5).

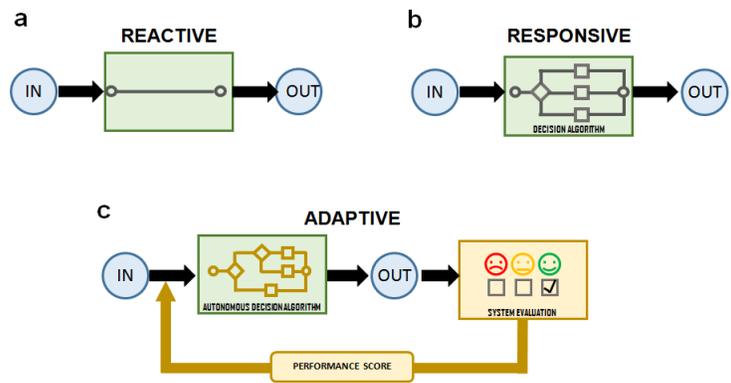

**Figure 4 – I/O systems and different 'intelligence' levels.**
**(a)** The reactive I/O system is capable of reading an input signal, but cannot interpret its meaning. The system's output is predefined by the human, based on theoretical assumptions or on empirical trial-and-error refinement. **(b)** The responsive I/O system is provided with a number of choices, but their conditional application is predefined by the human based on previously acquired knowledge. **(c)** The adaptive I/O system can independently choose the best output provided a varying input. The system learns the best output strategy through the feedback provided by the performance evaluator. The system may evolve and deliver a different output upon subsequent presentation of the same input based on the learned strategy and on its past experience.

## A simple decisional algorithm

In order to aid in the choice of the appropriate definition of (intelligent) biohybrid systems, we provide a simple decisional algorithm based on architecture and feature criteria (Fig. 5). This user-friendly flow-chart identifies three classes of operating I/O logic, from the basic non-intelligent types (*reactive* and *responsive*) to the most sophisticated self-evolving intelligent algorithm (*adaptive*).

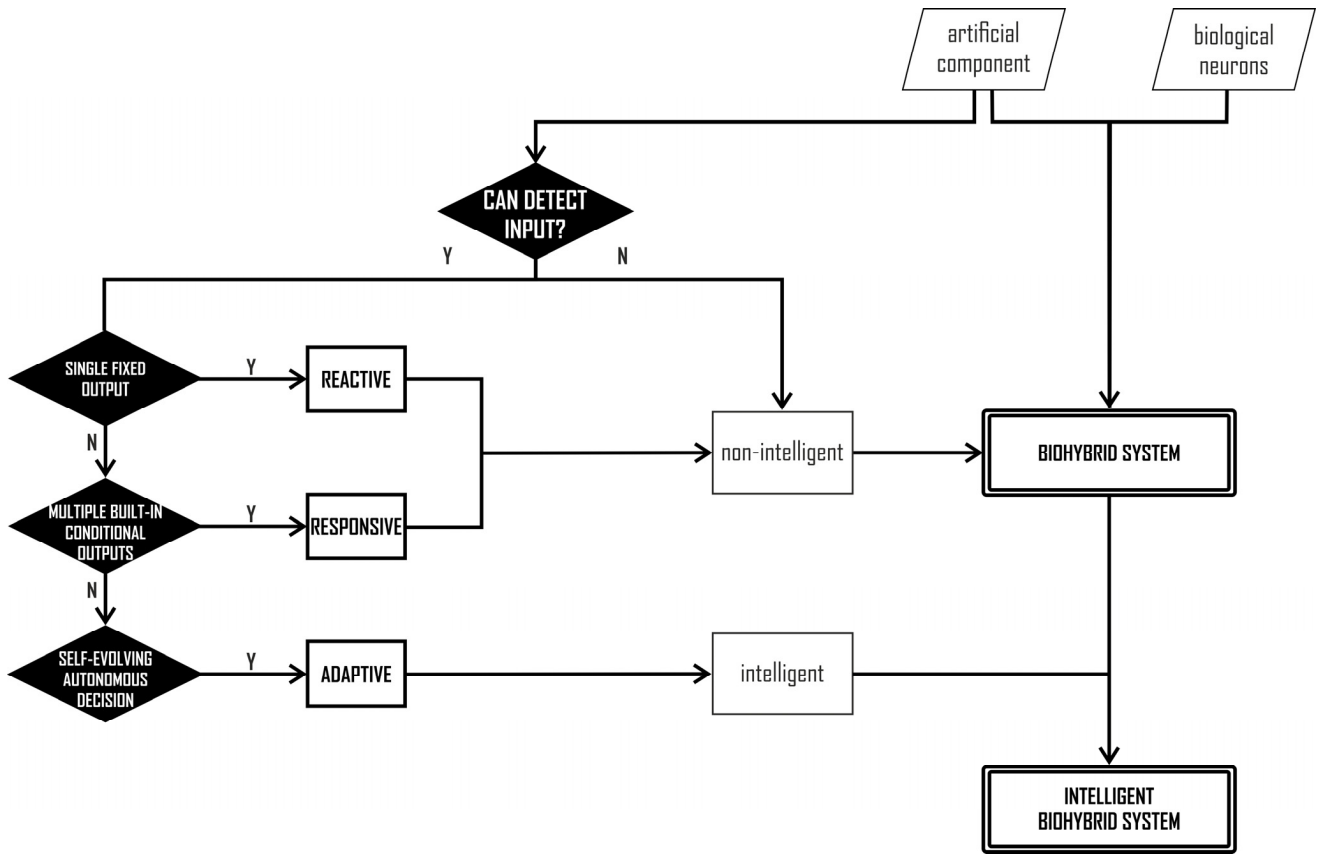

**Figure 5 – Is it biohybrid *and* intelligent?**

## Back to the future or back to the brain?

We have brought about the issues related to misused terminology in the emerging field of intelligent biohybrid neurotechnologies and we have proposed definitions to aid in the appropriate classification of such intelligent devices in the attempt to establish a global dictionary and look-up table for biologists and engineers. We conclude with one last provocative consideration about the 'evolution' of intelligent biohybrid systems.

The long-going effort of humans to build 'thinking computers' and artificial 'replacement parts' for the human body has landed on a very fertile ground. Needless to say, the inherent limitations of artificial devices have dictated the need of coupling them to biological components (typically neurons) in order to leverage their intelligence and transfer it to the machine. The most striking example is represented by the *living machines*, i.e., artificial devices with a human (organic) component, to the purpose of healing human diseases with artificial replacement parts[22]. More recently, there has been much excitement about 'neuronic computers', where unprecedented computational power is believed to be at hand by using neurons instead of transistors[23, 24]. This is a particularly intriguing paradox, since, in theory, machines should be there to aid humans, not the other way around. It is tempting to admit that despite the outstanding evolution of intelligent biohybrid neurotechnologies somehow we are back to the origins and that the provocation of Lady Ada Lovelace might have been meaningfully predictive: "Neurons think. Machines do". As progress in the neurotechnologies advances towards increasingly sophisticated 'intelligent' machines where biological neurons are leveraged as 'biological transistors', we dare raising a bigger dilemma: "Thinking computers or computing neurons?".

Humans have been striving for decades to mimic the brain using computing machines and have achieved sometimes remarkable goals (we refer in particular to the recent success of deep-convolutional neural networks in computer vision[25]). Still, it is clear that machines may be successful at solving specific problems but are still far from matching the capabilities of the human brain. After numerous efforts, the community has indeed eventually come to the conclusion that to achieve superior computational power, computational efficiency and, in the end, *real* intelligence, real neurons must be exploited as a novel organic material to build intelligent (biohybrid) machines[23, 24]. Whether this is 'evolution' we let the posterity judge. After all, the brain is said to be the most powerful computer…

**Acknowledgements:** G.P. is supported by the Marie-Skłodovska Curie Individual Fellowship Re.B.Us, Grant Agreement n. 660689, funded by the European Union under the framework program Horizon2020. We thank Dr Joelle Pineau for her useful comments.

**Authors' contribution:** conceived the manuscript – GP; elaborated the core conceptual design – GP and MS. All authors have written and approved the final version of the manuscript. Vector graphics by GP and MS.